%% file: main.tex
\documentclass{bmvc2k}
\graphicspath{{images/}}
\input{macros}

\title{Frequency learning for structured CNN filters with Gaussian fractional derivatives}

\addauthor{Nikhil Saldanha,\\Silvia L. Pintea,\\Jan C. van Gemert,\\ Nergis Tomen}{}{1}

\addinstitution{
Computer Vision Lab, \\
Delft University of Technology,\\
Delft, Netherlands
}

\runninghead{\scriptsize Nikhil Saldanha \etal{}}{\scriptsize Frequency learning for structured CNN filters}

\def\etal{\emph{et al}\bmvaOneDot}

\begin{document}
\maketitle
\input{abstract}
\input{introduction}

\input{related}

\input{method}
\input{experiments}
\input{discussion}

\input{conclusion}

\bibliography{fracsrf}

\end{document}

%% file: macros.tex
\usepackage{comment} 
\usepackage{mathtools}
\usepackage{xcolor}

\usepackage{multirow} 
\usepackage{wrapfig}
\usepackage{amssymb}

\newcommand{\mytitle}[1]{\smallskip\noindent\textbf{#1}}

\newcommand{\Eq}[1]{Eq.~(\ref{eq:#1})}
\newcommand{\eq}[1]{\Eq{#1}}
\newcommand{\fig}[1]{Fig.~\ref{fig:#1}}
\newcommand{\tab}[1]{Table~\ref{tab:#1}}

\DeclarePairedDelimiter\ceil{\lceil}{\rceil}
\DeclarePairedDelimiter\floor{\lfloor}{\rfloor}

\usepackage{pifont}
\usepackage{caption}
\usepackage{tabularx} 
\newcolumntype{Y}{>{\centering\arraybackslash}X}

\usepackage{booktabs}
\usepackage{multirow}

\newcommand{\ourModelName}{FracSRF\xspace}

%% file: abstract.tex

\vspace{-15px}

\begin{abstract}
Frequency information lies at the base of discriminating between textures, and therefore between different objects.
Classical CNN architectures limit the frequency learning through fixed filter sizes, and lack a way of explicitly controlling it. 
Here, we build on the structured receptive field filters with Gaussian derivative basis. 
Yet, rather than using predetermined derivative orders, which typically result in fixed frequency responses for the basis functions, we learn these.
We show that by learning the order of the basis we can accurately learn the frequency of the filters, and hence adapt to the optimal frequencies for the underlying learning task.
We investigate the well-founded mathematical formulation of fractional derivatives to adapt the filter frequencies during training. 
Our formulation leads to parameter savings and data efficiency when compared to the standard CNNs and the Gaussian derivative CNN filter networks that we build upon. 
\end{abstract}

%% file: introduction.tex
\section{Introduction}
\label{sec:intro}

\begin{wrapfigure}{r}{0.5\textwidth}
    \vspace{-15px}
     \includegraphics[width=0.5\textwidth]{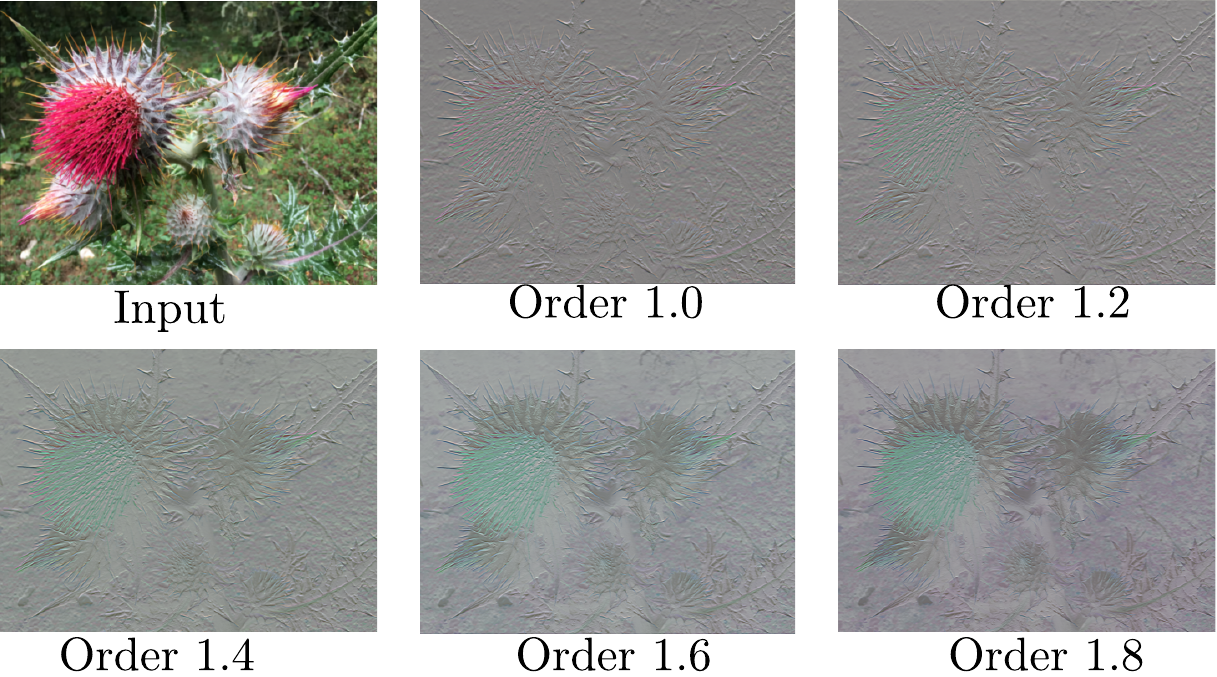}\\
    \caption{\small
    Filter responses when using fractional order Gaussian derivative filters (here x-order and y-order are equal).
    Defining the filters using fractional derivative orders adds flexibility in terms of the peak response frequency, and enables the use of standard gradient backpropagation for training.
    }
    \label{fig:motiv}
    \vspace{-10px}
\end{wrapfigure}

The world comes in many frequencies, and we rely on frequency as encoded in texture to differentiate between different object types: a purple thistle flower versus a purple tulip flower.
What\rq s more, convolutional neural networks (CNNs) additionally use texture (e.g. `fur' versus `skin') for discriminating between dissimilar object categories \cite{geirhos2018imagenet}.
Therefore, CNNs can reap benefits from an explicit lever for controlling the frequencies extracted from the data.

Current acclaimed CNNs architectures \cite{he2016deep,simonyan2014very,szegedy2016rethinking,tan2019efficientnet} lack an explicit knob to control the frequencies learned from the data. 
These classical CNN architectures hard-code the filter sizes thus limiting the frequency resolution contained in the filters.
Moreover, they learn each filter value separately at each featuremap location by treating the weights as independent, leading to data inefficiency. 
Here, we address both these issues, by proposing a way to explicitly control the frequency learning in a data-efficient continuous formulation using structured receptive fields with Gaussian basis.

We make the observation that the order of the Gaussian basis in the structured receptive fields (SRFs) \cite{Jacobsen2016CVPR} explicitly controls the maximum frequency of the filters, and therefore the maximum frequencies they can detect in the data. 
We, additionally, observe that when using SRFs~\cite{Jacobsen2016CVPR}, typically a few Gaussian basis functions are sufficient to extract useful information.
However, while it may be adequate to use a single basis function out of the whole basis to define each kernel, selecting from a large range of derivative orders may be necessary.
Putting together these observations, we aim to learn a single Gaussian derivative per kernel where the order of the Gaussian derivative is adapted during training to better represent the frequencies present in the data.
Typically, the derivative order is an integer (e.g. first order derivative or second order derivative) which makes backpropagation difficult. However, the order of the Gaussian derivatives become differentiable when working within the domain of fractional calculus. In this work, we make use of the fractional derivatives of the Gaussian function to learn the derivative order.
\fig{motiv} shows examples of image responses when using fractional order Gaussian derivatives. 
Fractional orders add flexibility in terms of the frequencies that the model can encode and make the model easily trainable using standard gradient backpropagation methods. 

This article makes the following contributions: 
(i) We propose a well-founded method for learning the filter frequencies from data, and demonstrate its effectiveness experimentally;
(ii) To that end, we describe a mathematically solid approach to learning fractional order Gaussian derivatives; 
(iii) We demonstrate improved data efficiency and parameter savings across 4 datasets when comparing with existing standard CNNs and baselines with structured CNN filters.

%% file: related.tex
\section{Related Work}
\mytitle{Structured filters in CNNs.}
Influential prior work has investigated the usefulness of structured filters for image analysis.
Simoncelli \etal~\cite{simoncelli1992shiftable} define a steerbale pyramid using a set of wavelets that encode orientation and scale, while Mallat defines complex wavelet basis filters in ~\cite{mallat1999wavelet}. 
These complex wavelets have been used in the Scattering transform~\cite{brunaPAMI13scatter,mallat2012groupInvScat} which is later extended in ~\cite{cotterMLSP17visualizing,oyallonICCV17scalingScattering,sifreCVPR13rotScat,singhCVPRw17efficientScatter}.
Other works consider PCA basis ~\cite{ghiasiECCV16laplacian}, Gabors~\cite{luan2018gabor,perez2020gabor}, circular harmonics~\cite{WorrallCVPR17harmonicNets}, or simply learning the basis from the data \cite{li2019learning}. 
A large amount of work has been focused on Gaussian derivatives basis~\cite{Jacobsen2016CVPR} used for controlling the scale in deep networks~\cite{sosnovik2019scale,lindeberg2020scalecovariant,pintea2021resolution} or for making the networks continuous over space and depth~\cite{tomen2021deep}. 
Here, we also build on the Gaussian derivative basis~\cite{Jacobsen2016CVPR} because it allows us to easily control the number of learnable parameters by directly learning the order of the Gaussian derivative basis.
The order parameter controls the complexity of the patterns the filters can respond to, therefore by learning the order we learn how complex these filters need to be.
While wavelets, such as Gabor filters, can directly learn the frequency response of the filters, the frequency parameter of the wavelet is coupled to its scale which relates to its spatial extent. 
Our representation decouples the frequency response and the scale/spatial extent of the filters, via two independently trained parameters: derivative order and scale-parameter $\sigma$.

\mytitle{Parameter efficiency and data efficiency in CNNs.}
CNNs come with large computational costs entailed by the large number of parameters to be learned on the training data. 
A new trend is emerging with focus on efficiency.  
Model compression has been the most intuitive manner of reducing computations and memory~\cite{han2015deep,he2018amc,yang2018netadapt}.
Alternatively, the use of $1\times1$ convolutions have significantly reduced the parameters in SqueezeNets~\cite{iandola2016squeezenet,gholami2018squeezenext}.
Depthwise separable convolutions combined with $1\times 1$ convolutions have shown parameter efficiency~\cite{chollet2017xception,howard2017mobilenets,ma2018shufflenet,zhang2018shufflenet}. 
More recently EfficientNet~\cite{tan2019efficientnet} shows both accuracy improvement and parameter reduction by carefully scaling network width, depth and resolution.
Similarly, here we also propose a model aimed at reduced parameters by learning how complex the filters need to be. 
Moreover, our proposed fractional structured filters can be used in combination with any efficient convolutional architecture. 


    
\mytitle{Frequency learning in CNNs.}
Analyzing the deep networks in frequency domain has brought insights into how they work.
Deep networks can fit, barely perceivable, high-frequency signals, thus leading to vulnerability to adversarial attacks \cite{wang2020high,yin2019fourier,tomen2021spectral}. 
However they tend to learn low frequency signals first \cite{rahaman2019spectral}.
Rather than using frequency domain to analyze deep networks, the networks can actually be trained in the frequency domain~\cite{goldberg2020rethinking,watanabe2020image} or over inputs transformed to the frequency domain~\cite{xu2020learning}.
Here, we also analyze which frequencies our model can fit well and where it makes errors. 
Our proposal learns the appropriate frequency of the filters by learning the order of the Gaussian basis.

%% file: method.tex
\section{Fractional structured filters}
\subsection{Review of Gaussian basis filters}
Rather than representing filters as a discrete set of pixel values, the use of Scale-space theory~\cite{lindeberg_scale-space_1994,witkin_scale-space_nodate} enables the definition of filters as continuous functions \cite{Jacobsen2016CVPR,sosnovik2019scale,tomen2021deep}.
And instead of learning the values of the individual pixels, one only needs to learn the parameters of these functions.
The underlying idea is that a filter $F(x)$ can be approximated with a Taylor expansion around a point $a$, up to an order $N$:
\begin{equation}
    F(x) \approx \sum_{i=0}^{N} \frac{F^i(a)}{i!}(x-a)^i.
\end{equation}
Scale-space theory ~\cite{lindeberg_scale-space_1994,witkin_scale-space_nodate} defines the filter derivatives $F^i$ as the convolution ($*$) of the filter $F$ with Gaussian derivatives, $G^i$:
\begin{equation}
    F(x) \approx \sum_{i=0}^{N}\frac{(G^i(.;\sigma)*F)(a)}{i!}(x-a)^i
    \label{eq:taylor2}
\end{equation}
where $\sigma$ is the standard deviation of the Gaussian representing the scale parameter \cite{lindeberg_scale-space_1994}.     
The recursive formulation relying on Hermite polynomials \cite{martens1990hermite} allows to effectively compute the $i^{th}$ Gaussian derivative $G^i$ as a point-wise multiplication ($\circ$) between the Gaussian $G$ and the $i^{th}$ Hermite polynomial, \(H_i\):
    \begin{equation}
        G^i(x;\sigma) = \left(\frac{-1}{\sigma\sqrt{2}}\right)^i H_i\left(\frac{x}{\sigma\sqrt{2}}\right)\circ G(x;\sigma),
    \end{equation}
where the recursive definition of the Hermite polynomials is:  $H_0(x) = 1$; $H_1(x) = 2x$; $H_i(x) = 2x H_{i-1}(x) - 2(i-1)H_{i-2}(x)$. 

By simplifying \eq{taylor2} and incorporating the polynomial coefficients in a set of weights $\alpha$, previous work \cite{Jacobsen2016CVPR,tomen2021deep} defines the filter approximation $F$ as a linear combination of Gaussian derivatives up to order $N$:
\begin{equation}
    F(x, \sigma) \approx \sum_{i = 0}^{N} \alpha_{i} G^{i}(x;\sigma),
    \label{eq:srf}
\end{equation}
where both the weights $\alpha$ and the scale parameter $\sigma$ are can be learned from data \cite{pintea2021resolution,tomen2021deep}.

\subsection{Fractional structured filters: Learning the basis order}
    
\begin{figure}[t]
    \begin{tabular}{c}
    \includegraphics[width=0.95\textwidth]{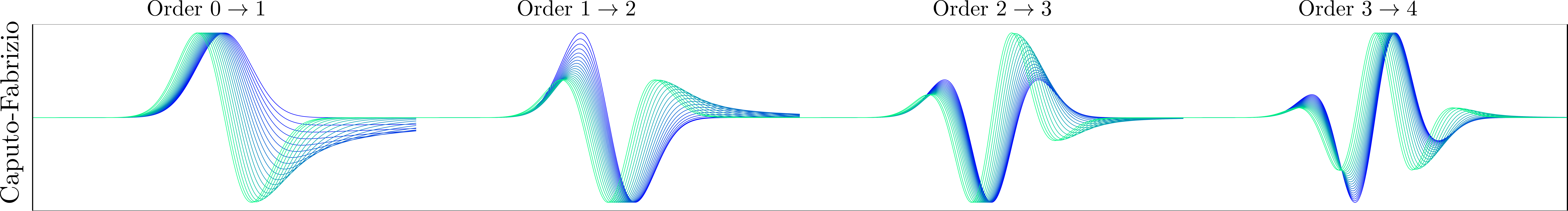}\\
    \includegraphics[width=0.95\textwidth]{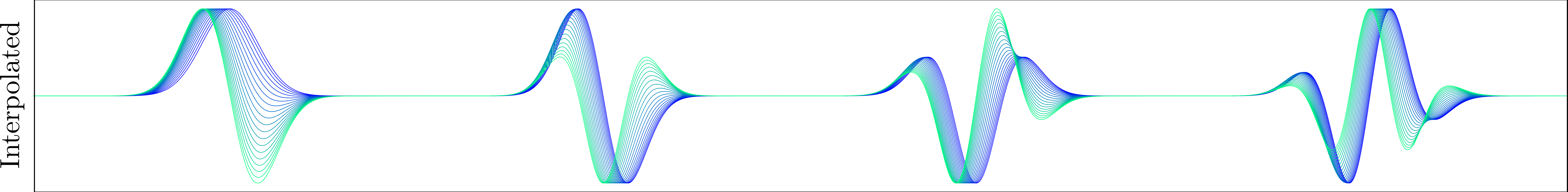}\\[5px]
    \end{tabular} 
    \caption{\small
        Top: Gaussian derivatives computed using Caputo-Fabrizio \cite{caputo2015new} fractional derivative form.
        Bottom: Fractional Gaussian derivatives computed via interpolation between integer orders.
        The error introduced by using the interpolation is small relative to the Caputo-Fabrizio form.
        } 
    \label{fig:interp_vs_cf_derivative} 
    \vspace{-10px}
\end{figure} 

We propose to learn the frequency of the filters by making the order of the Gaussian basis a learnable parameter.
Instead of defining the filter as a linear combination of Gaussian derivatives up to order $N$, as previously done \cite{Jacobsen2016CVPR}, we approximate the filter with only one weighted Gaussian derivative, where the order of the derivative $\nu$ is a learnable parameter: 
\begin{equation}
        F(x;\sigma) \approx \alpha G^{\nu}(x; \sigma).
        \label{eq:frac_srf}
\end{equation}
When using this filter definition in a deep network, we can obtain the gradients of the loss function with respect to $\nu$ through the standard network backpropagation.
One caveat of learning the order of the Gaussian derivative is that a gradient descent step will always result in real (fractional) order updates. 
Since the Gaussian derivatives are traditionally only defined for integer orders, we need to account for orders in between two integers. 

One possible way of dealing with fractional derivatives is the Caputo-Fabrizio \cite{caputo2015new} form, which in the 1D case is:  
\begin{equation}
    G_{CF}^{\nu}(x; \sigma) = \frac{1}{1-\nu}\cdot\frac{1}{\sqrt{2\pi}\sigma^3}\exp{\left(-\frac{1}{1-\nu}\left(x  - \frac{\sigma^2}{2}\cdot\frac{1}{1-\nu}\right)\right)}\zeta_{\sigma, \nu}(x)
    \label{eq:cf_deriv}
\end{equation}
where $\zeta_{\sigma, \nu}(x)$ is an integral of the form:
\begin{equation}
    \zeta_{\sigma, \nu}(x) = \int_0^t (\mu-x) \exp{\left(-\frac{\left(\tau + \frac{1}{1-\nu}\sigma^2\right)^2}{2\sigma^2}\right)}d\tau
\end{equation}
However, when using this formulation, we observed exploding gradients due to the non-linear terms. 
A more straight-forward approach is to interpolate between the two closest integers of the fractional order: 
\begin{equation}
    G_{Iter}^\nu (x; \sigma) = (\ceil*{\nu} - \nu)  G^{\floor*{\nu}} (x; \sigma) + (\nu - \floor*{\nu}) G^{\ceil*{\nu}}(x; \sigma), 
    \label{eq:inter_deriv}
\end{equation}
where $\ceil{\cdot}$ and $\floor{\cdot}$ are the ceil and floor roundings of $\nu$.
This formulation permits us to keep the gradients in check, due to linear nature of interpolation used.
\fig{interp_vs_cf_derivative} shows a number of fractional order Gaussian derivatives when going from order 0 to 1, 1 to 2, 2 to 3, and 3 to 4. On the top row the Caputo-Fabrizio form (\eq{cf_deriv}) is used for computing the 1D derivatives, while on the bottom row the interpolation method (\eq{inter_deriv}) for estimating fractional Gaussian derivatives. There is on average less than 0.22 root mean squared error between these two estimations.
In all our experiments we use the linearly interpolation method to compute the fractional Gaussian derivatives.

Because we are working with images, we use 2D Gaussian derivatives. The outer product ($\otimes$) of 1D Gaussian derivatives along the $x$- and $y$-direction defines the 2D Gaussian derivative: \(G^{i+j}(x,y;\sigma) = G^{i}(x;\sigma)\otimes G^{j}(y;\sigma) \).

\subsection{Deep networks with fractional structured filters}
\begin{figure}[t]
    \centering
    \small
    \begin{tabular}{@{}ccc@{}}
    \includegraphics[width=0.29\textwidth]{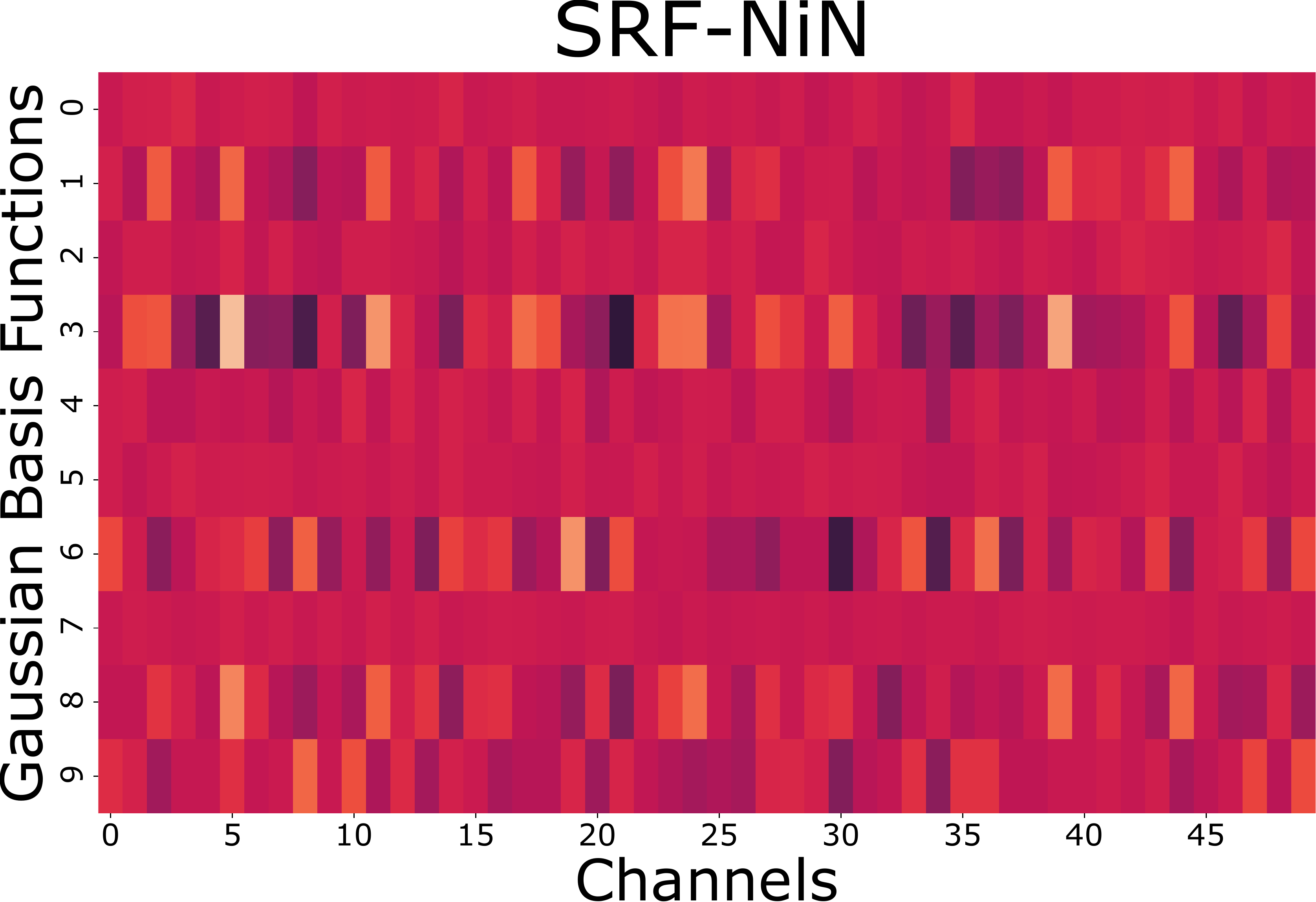} &
    \includegraphics[width=0.32\textwidth]{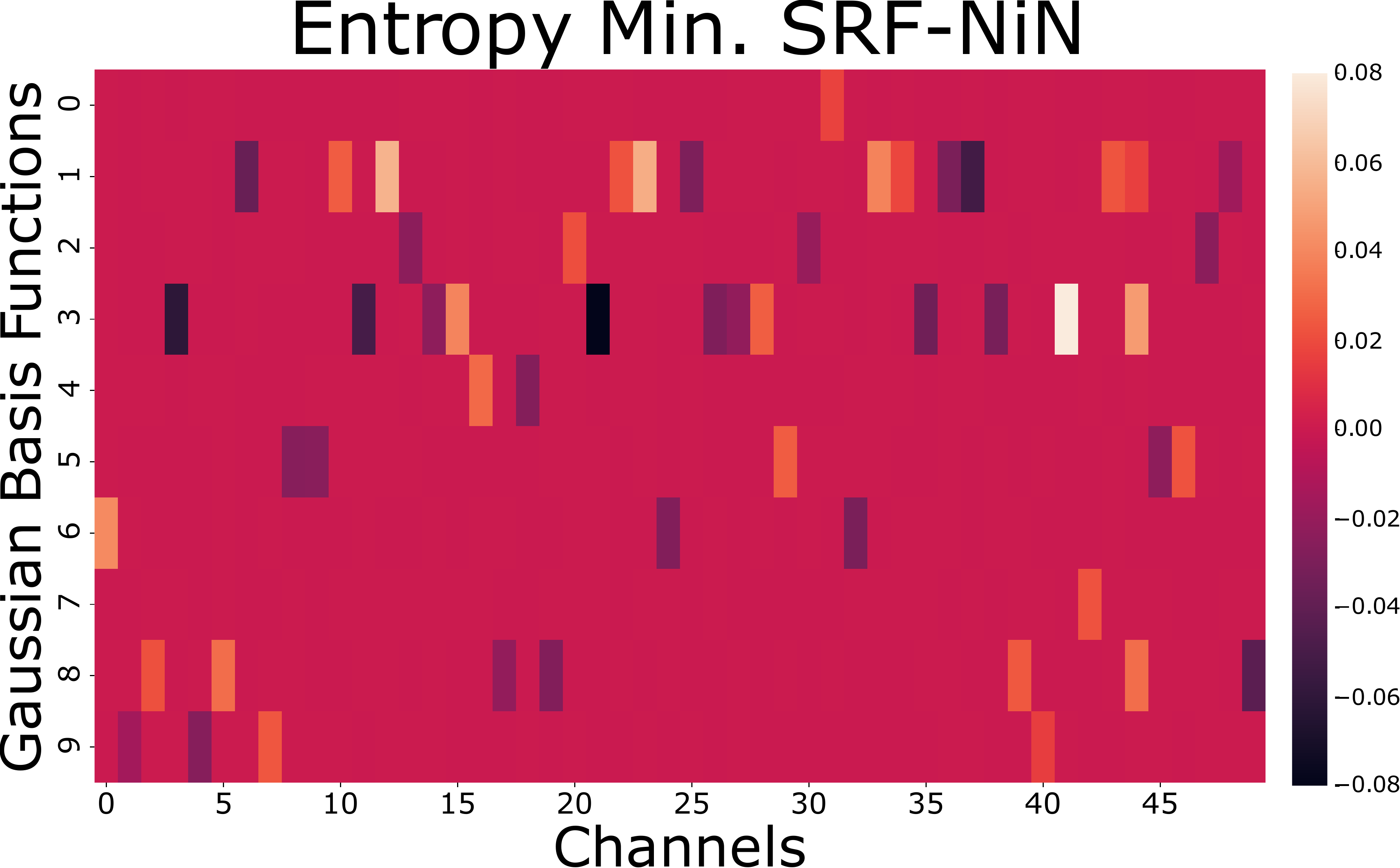} &
    \includegraphics[width=0.31\textwidth]{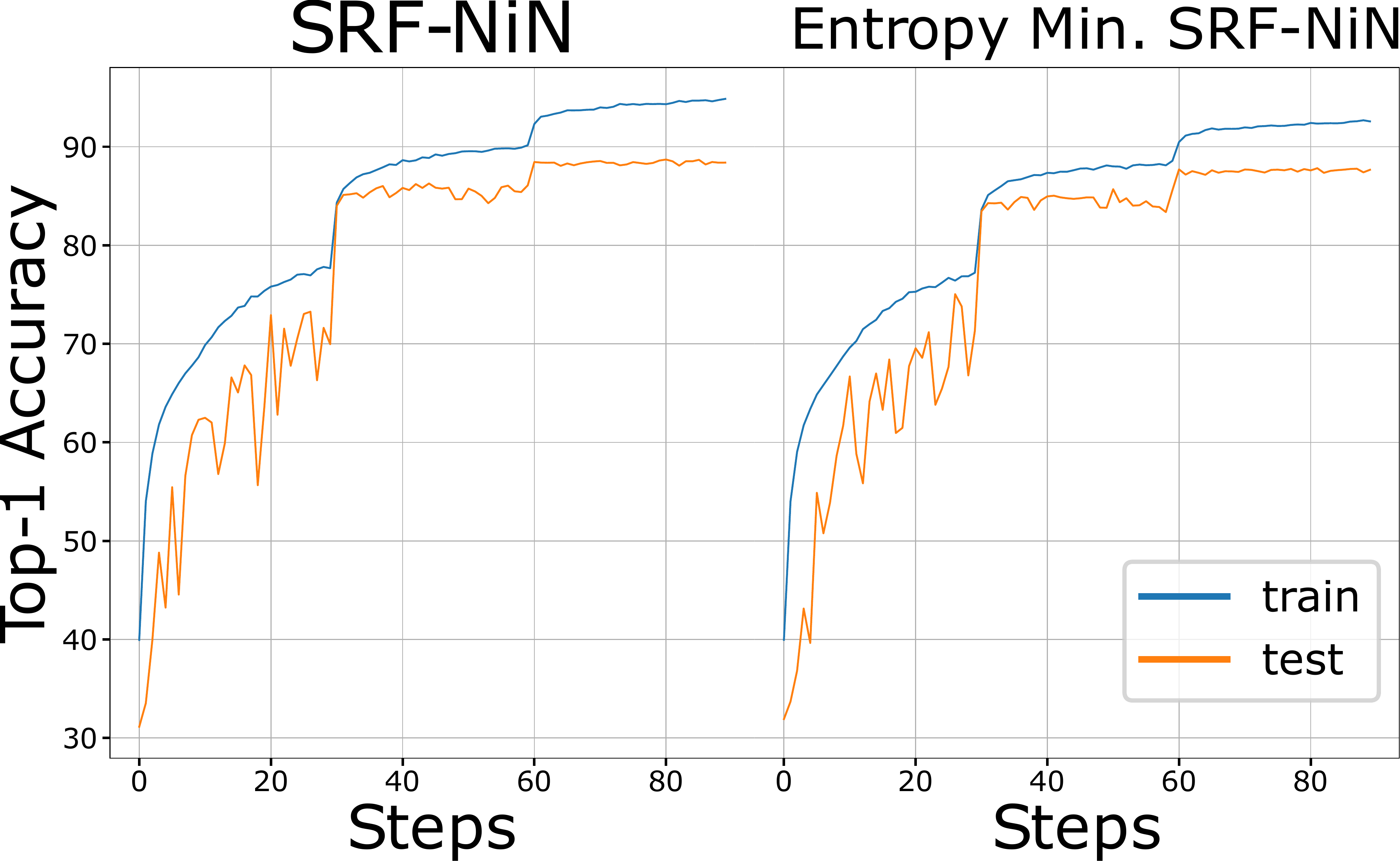}\\
    (a) & (b) & (c)\\[3px]
    \end{tabular}
    \caption{\small
        (a) The distribution of $\alpha$ weights (color bar) learned in a layer of the original SRF-NiN \cite{Jacobsen2016CVPR} model on \emph{CIFAR-10}. 
        (b) The distribution of $\alpha$-s when minimizing their entropy.
        (c) Training/test accuracies for the original SRF-NiN and the entropy-minimized version.  
        We can safely reduce the number of Gaussian derivatives defining the filters (i.e. set most basis coefficients $\alpha$ to zero), at no cost to validation accuracy.
    } 
    \label{fig:entropy} 
    \vspace{-10px}
\end{figure} 

Each 2D Gaussian derivative requires two order parameters: the order on the x-axis, $\nu_x$, and the order on the y-axis, $\nu_y$. 
When considering a filter $F$ of size $[C,K,W,H]$ with $C$ input channels and $K$ output channels, we learn in practice two order parameters ($\nu^x_{ck}$, $\nu^y_{ck}$) per kernel in the filter, and a scalar ($\alpha_{ck}$) for each kernel:
\begin{equation}
        F(c,k,x,y; \sigma) = \alpha_{ck} ( G^{\nu^x_{ck}}(x;\sigma)\otimes G^{\nu^y_{ck}}(y;\sigma) ), 
\end{equation}
where the scale parameter $\sigma$ is shared among the kernels in the filter and can either be learned as in \cite{pintea2021resolution,tomen2021deep}, or fixed as in \cite{Jacobsen2016CVPR,sosnovik2019scale}. 
Our method is more flexible than the structured receptive fields (SRF) \cite{Jacobsen2016CVPR}, allowing for non-integer derivatives. 
We coin our filters \ourModelName.

\mytitle{Is one Gaussian derivative sufficient?} Unlike previous work \cite{Jacobsen2016CVPR,tomen2021deep}, we do not use a linear combination of Gaussian derivatives up to a fixed order. 
We use a single Gaussian derivative, whose order can be learned.  
To check whether using a single Gaussian derivative is sufficient, we do a small test on the \emph{CIFAR-10} dataset, using SRF filters \cite{Jacobsen2016CVPR} over a NiN \cite{lin2013network} backbone. 
In the SRF model the $\alpha$ weights control how much a certain integer-order Gaussian derivative contributes to the final filter.  
\fig{entropy}.(a) shows the distribution of the $\alpha$-s in a layer of the original SRF-NiN model, compared to the same model in \fig{entropy}.(b) where we normalize the $\alpha$ values and we minimize their entropy.
Minimizing the entropy of $\alpha$-s reduces the actual number of Gaussian derivatives used per filter.
At no loss in accuracy (\fig{entropy}.(c)) the number of Gaussian derivatives can be reduced from 9 to 2 per channel. 
This supports our intuition that using one Gaussian derivative is sufficient, where we make it more flexible by learning its order from the data.

%% file: experiments.tex
\section{Experiments}

\subsection{Experimental setup}

\mytitle{Datasets.} We test our method across 4 datasets: \emph{CIFAR-10}, \emph{CIFAR-100} \cite{krizhevsky2009learning}, and \emph{STL-10} \cite{coates2011analysis} and ImageNette \cite{howard2020fastai}, having low and high resolution images, respectively. 
Additionally, to test the method\rq s ability to learn the correct data frequency, we created a dataset called \emph{Sinusoids} containing 2D sinusoids of various orientations and 5 spatial frequencies defining the 5 classes. 
We also test our method\rq s accuracy in few-data samples regime by sub-sampling the \emph{CIFAR-10} dataset between 40 and 0.04\% of the original number of images.

\mytitle{Models.} We consider several backbone architectures: Network in Network (NiN) \cite{lin2013network}, Resnet-32 \cite{he2016deep}, EfficientNet-b0 \cite{tan2019efficientnet}. 
We also compare with a few methods using structured filters: SRF \cite{Jacobsen2016CVPR, pintea2021resolution}. 
To obtain the SRF and our \ourModelName variants, we replace all the non 1$\times$1 convolutional layers either with SRF layers or with \ourModelName layers. 
For the SRF networks, we always set the Gaussian basis orders to 2.
For our models we initialization of the orders uniformly between $[1, 6]$, set the spatial filter extent to $2\sigma$ around the center and initialize $\sigma=1$, unless stated otherwise.
We train using SGD with momentum of 0.9 and $L_2$ regularization of 5e-4.
For \ourModelName-NiN, \ourModelName-Resnet32, \ourModelName-Efficientnetb0 we use a learning rate of 0.1,
0.05, 0.001 and batch sizes of 128, 256, and 16.
When enabling $\sigma$ learning in \ourModelName, we use a different learning rate and weight decay for $\sigma$ of 0.001 and 0.01 on \ourModelName-Resnet-32, while on \ourModelName-EfficientNet-b0 we use 0.001 and 0.05 for $\sigma$ learning. 
We keep learning rates and batch sizes fixed across datasets except for \ourModelName-Efficientnet-b0 on \emph{STL-10} where due to memory limitations, we use a batch size of 4 and learning rate proportionally increased to 0.05.
For the baselines NiN, Resnet-32 and EfficientNet-b0 we use learning rates of 0.1, 0.01, 0.01 and batch sizes of 128, 128 and 16, respectively. 
Given the relatively small dataset sizes, we use the lightweight version of \emph{Resnet-32} where the first block has 16 channels and the last block 64.
For the SRF-NiN, SRF-Resnet-32 and SRF-EfficientNet-b0 we use learning rates of of 0.1, 0.05, 0.001 and batch sizes of 128, 256, and 16 respectively.

\begin{figure}[t]
    \centering
    \small
    \begin{tabular}{c@{\hskip 0.3in} c} 
    \includegraphics[width=0.4\textwidth]{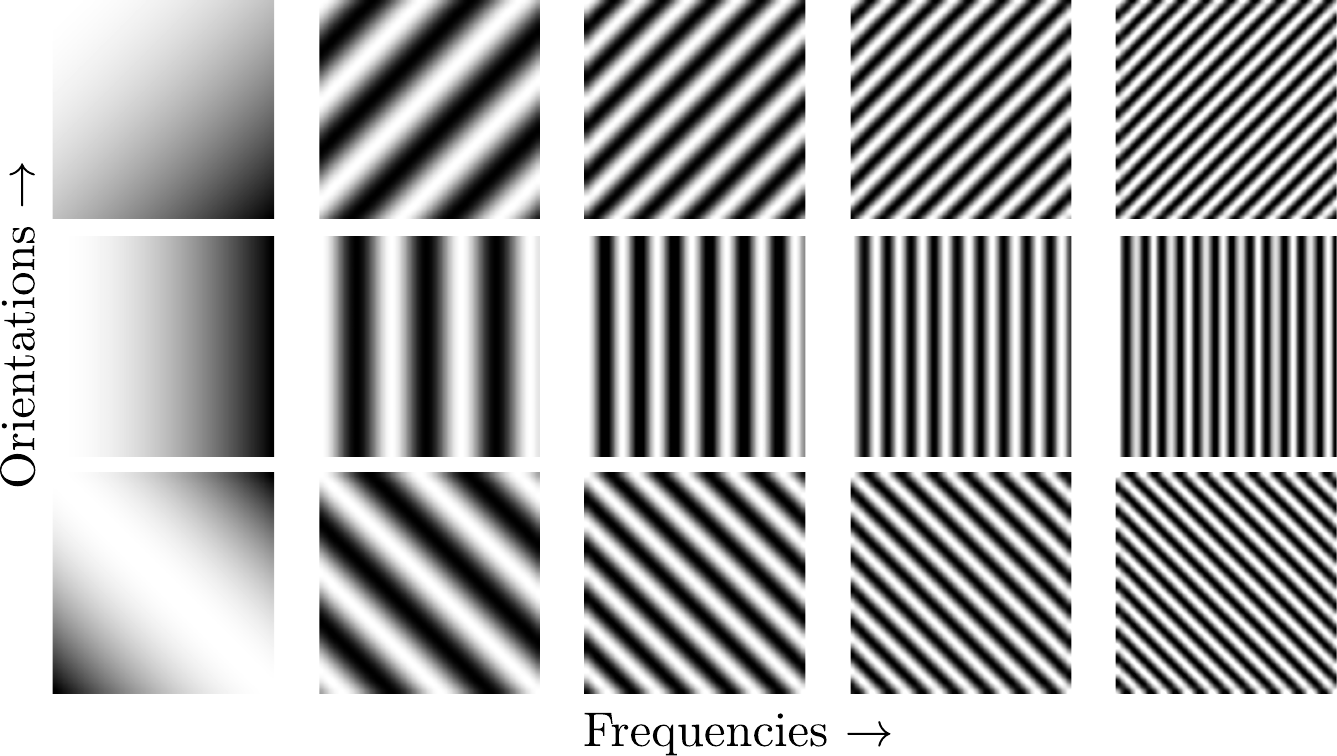} & 
    \includegraphics[width=0.4\textwidth]{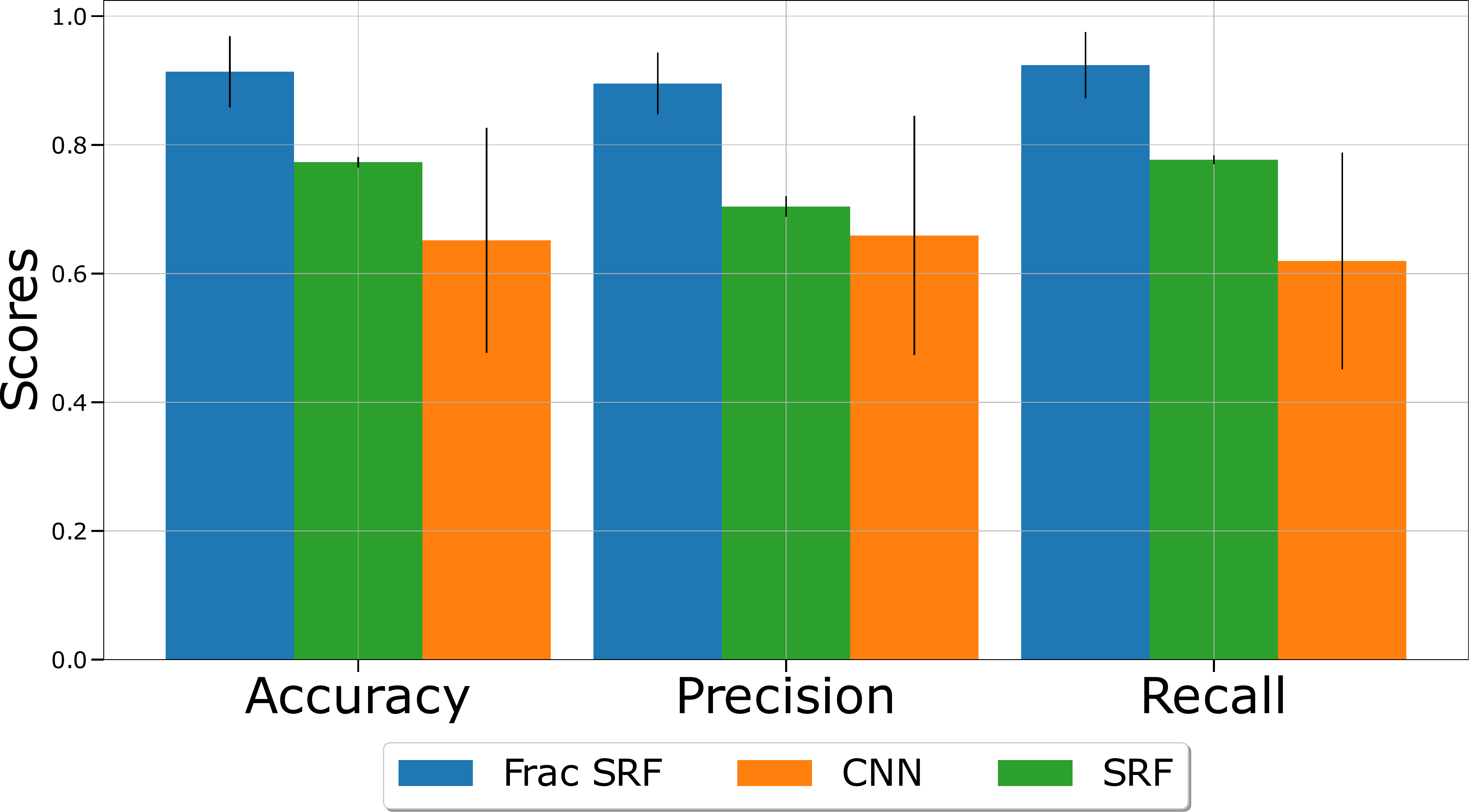}\\
    (a) \emph{Sinusoids} dataset & (b) \emph{Sinusoids} scores \\[3px]
     \end{tabular} 
    \caption{ \small \textbf{Exp 1:}
    (a) Examples from the toy \emph{Sinusoids} dataset. We vary the number of frequencies and the orientations.  
    (b) Accuracy / Precision / Recall results on the \emph{Sinusoids} dataset. For a baseline CNN, its SRF equivalent, and \ourModelName. 
    Our \ourModelName is more suitable for learning varying frequencies.
    }
    \label{fig:sinusoids}
\end{figure}
\begin{figure}[t]
    \centering
    \begin{tabular}{c@{\hskip 0.4in} c@{\hskip 0.4in} c} 
    \includegraphics[width=0.23\textwidth]{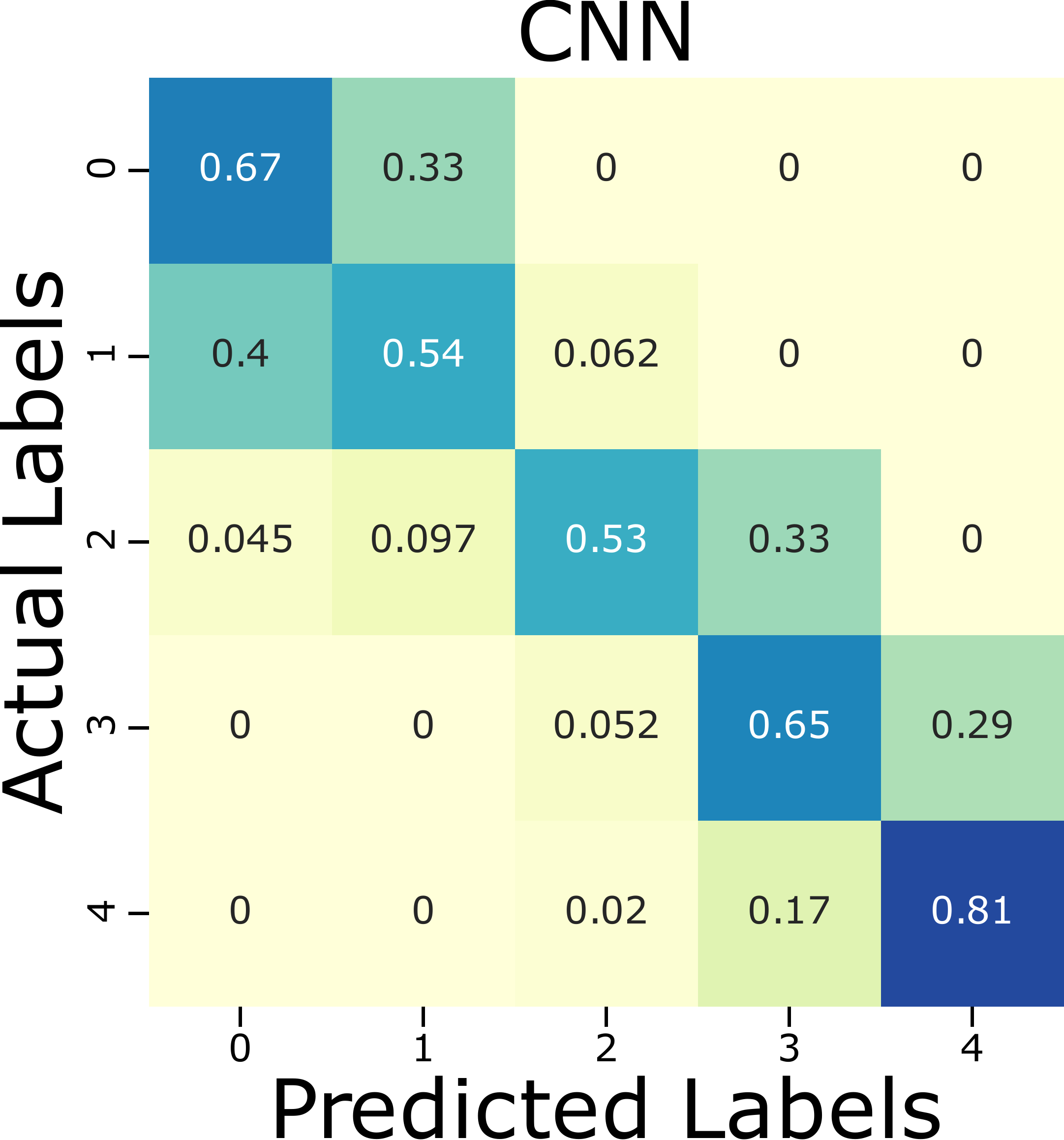} & \includegraphics[width=0.23\textwidth]{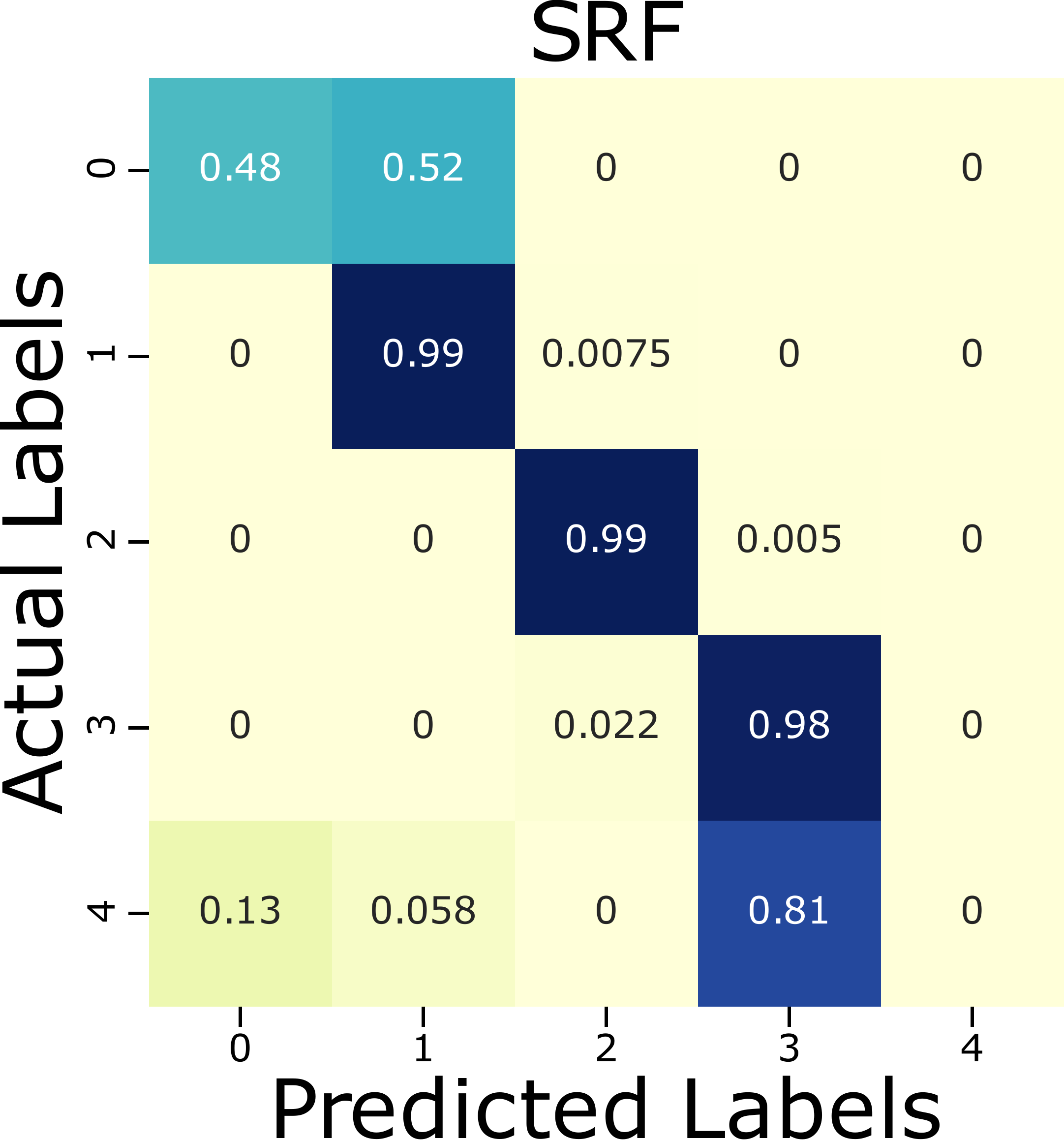} &
    \includegraphics[width=0.26\textwidth]{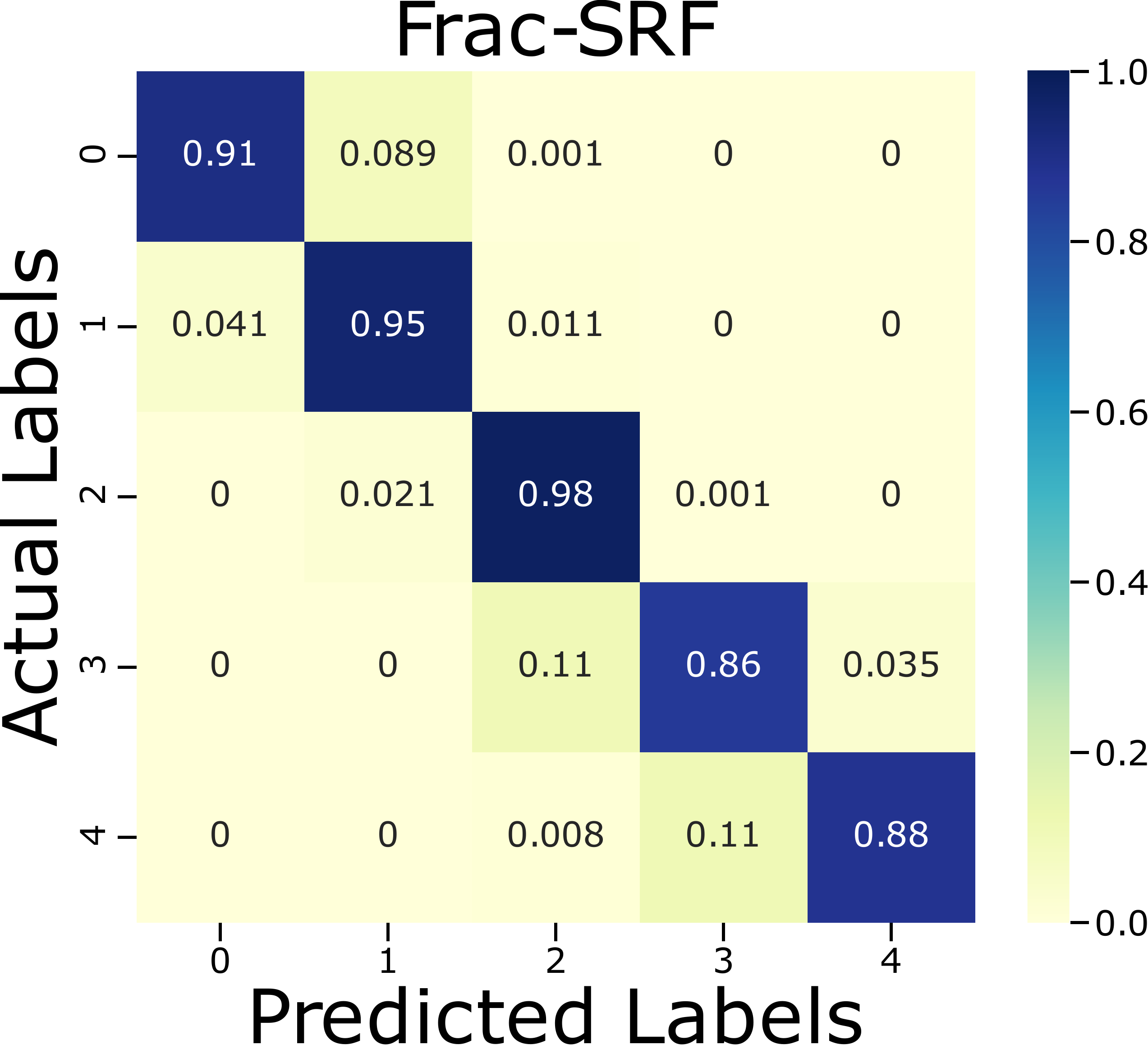}\\[5px] \end{tabular} 
    \caption{\small \textbf{Exp 1:}
    Confusion matrices for the CNN, SRF \cite{Jacobsen2016CVPR}, and \ourModelName small networks on the \emph{Sinusoids} dataset.
    Our \ourModelName can learn varying frequencies, and therefore it is better at distinguishing the 5 classes.
    }
    \label{fig:confs}
    \vspace{-10px}
\end{figure} 

\subsection{\emph{Exp 1:} Does \ourModelName learn the correct data frequency?}
We test the hypothesis that our \ourModelName is more flexible in learning a large range of frequencies, by learning the Gaussian derivative order. 
For this we create a synthetic toy dataset coined the \emph{Sinusoids} dataset.
\fig{sinusoids}.(a) shows a few examples from this dataset. 
The dataset contains 5 classes, each with 600 training examples and 200 test examples.
Each class corresponds to a different frequency, where we vary the orientations of the sinusoids across examples.   
For this experiment we use a small 2-layer network where the first layer has 32 output channels and the second 5 output channels. 
We repeated the experiments $5 \times$.
For the normal CNN we learn the filters the traditional way, for the SRF we replace the filters with a linear combination of Gaussian derivatives as in \cite{Jacobsen2016CVPR} with $\sigma = 1$, and for \ourModelName we use a single weighted Gaussian derivative with $\sigma=1$. 
All filters are $5\times 5$ px.

\fig{confs} shows confusion matrices for the CNN, SRF \cite{Jacobsen2016CVPR} and \ourModelName 2-layer networks on the \emph{Sinusoids} dataset.
\fig{sinusoids}.(b) reports accuracy, precision and recall scores for these three methods.
SRF cannot predict the highest frequency classes, being limited by its fixed order in the Gaussian basis. 
The CNN is not able to resolve between similar frequencies and tends to confuse neighboring classes.
Our \ourModelName can learn the varying frequencies and therefore is able to better separate the 5 frequency classes. 

\begin{table}[t]
    \centering
    \resizebox{0.9\columnwidth}{!}{
    \begin{tabular}{@{}lccccl@{}}
    \toprule
    & \multicolumn{5}{c}{Filter scale initialization} \\ \cmidrule(l){2-6} 
     & $\sigma= 2^{-2}$ & $\sigma= 2^{-1}$ & $\sigma= 2^{-0}$ & $\sigma= 2^{1}$ & $\sigma= 2^{2}$ \\ \midrule
    Top-1 Accuracy (\%) & $90.59\pm0.2$ & $90.65\pm0.04$ & $90.90\pm0.05$ & $90.68\pm0.25$ & $90.26\pm0.29$  \\
    Initial Filter Size  & 3$\times$3 & 5$\times$5 & 7$\times$7 & 9$\times$9 & 11$\times$11 \\
    Training Time (sec/epoch)  & 79.2s & 79.2s & 79.8s & 79.2s & 81.0s \\
    \bottomrule
    \end{tabular}
    }
    \caption{\small \textbf{Exp 2.(a):}
        Impact of initializing the filter scale on the performance and training time of the \ourModelName-NiN on \emph{CIFAR-10}.
        The network can adapt the scale parameter $\sigma$ even when initialized far from the optimum.
        The best initialization seems to be $\sigma=2^0$.
    }
    \label{tab:scale_vs_top1}
    \vspace{-10px}
\end{table}

\begin{table}[t]
    \centering
    \resizebox{0.7\columnwidth}{!}{
    \begin{tabular}{@{}llcc@{}}
    \toprule
    & \multicolumn{3}{c}{Filter order initialization} \\ \cmidrule(l){2-4} 
    & order $\in \mathcal{U}_{[1,3]}$ & order $\in \mathcal{U}_{[3,6]}$  & order $\in \mathcal{U}_{[6,10]}$ \\ \midrule
    Top-1 Accuracy (\%) & $90.62\pm0.20$ & $90.34\pm0.12$ & $89.52\pm0.13$ \\
    Training Time (sec/epoch) & 74.4s & 74.5s & 79.2s \\
    \bottomrule
    \end{tabular}
    }
    \caption{\small \textbf{Exp 2.(b):}
    Impact of order initialization on CIFAR-10 using \ourModelName-NiN.
    There is not a large difference in performance between different order ranges used for initialization.
    The model can learn to adapt the order to the best one.
    Higher orders require more computations. 
    }
    \label{tab:order_vs_top1}
    \vspace{-10px}
\end{table}

\subsection{\emph{Exp 2:} Model choices analysis}
\mytitle{\emph{Exp 2.(a):} Impact of scale initialization.}
We test the effect of the initialization of the scale parameter ($\sigma$) of the Gaussian derivatives, in our \ourModelName filters. 
Following \cite{tomen2021deep} we learn the $\sigma$ and initialize it as a power of 2, which avoids dealing with negative $\sigma$ gradients during training.
And we initialize the order uniformly in $[1, 6]$.
\tab{scale_vs_top1} shows results across 3 repetitions when varying $\sigma$ for the \ourModelName-NiN on \emph{CIFAR-10}. 
The initialization of the scale parameter shows minors variations, with $\sigma=2^0$ being the best.
The network can correct for the scale well even when initialized far away from the optimum. 
Additionally, using larger scales impacts the training time.

\mytitle{\textbf{Exp 2.(b)}: Impact of order initialization.}
We test the effect of initializing the Gaussian derivative order on the \emph{CIFAR-10} dataset using \ourModelName-NiN. 
We vary the initialization of the order by uniformly sampling in the ranges: $[1, 3]$, $[3, 6]$, and $[6, 10]$. We repeated the experiments $3\times$.
\tab{order_vs_top1} shows the optimal order initialization is found in the interval $[1, 3]$.
There is not a large difference between the different initialization ranges, suggesting that the model can learn the correct orders for task. 
Starting from larger order range is sub-optimal as the training time increases: the Hermite polynomial computations requires more time at higher orders. 
\emph{CIFAR-10} does not contain many high frequencies and therefore it is reasonable that orders up to 3 are able to capture the information.

\begin{figure}[t]
    \centering
    \begin{tabular}{c} 
        \includegraphics[width=0.9\textwidth]{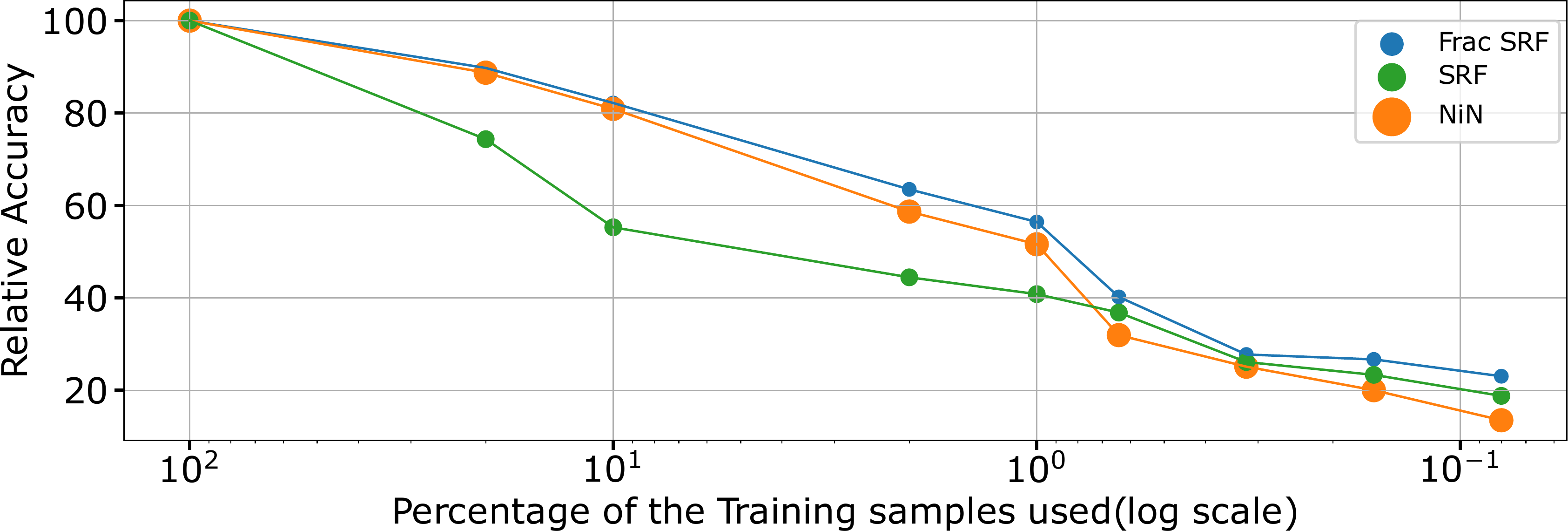}\\
    \end{tabular}
    \vspace{0.1cm}
    \caption{\small \textbf{Exp 3.(a):}
     Data efficiency in the FracSRF model. Relative accuracy of NiN, SRF-NiN \cite{Jacobsen2016CVPR, pintea2021resolution}, 
    and \ourModelName-NiN on subsets of \emph{CIFAR-10}. 
    The dot size of each method indicates the relative number of parameters.
    The performance of each model are normalized as a percentage of their own accuracy at 100\% training data.
    The scores of our \ourModelName-NiN degrade less rapidly especially when compared to NiN and SRF-NiN.
    }
    \label{fig:performance_subset_exp}
    \vspace{-10px}
\end{figure}
\subsection{\emph{Exp 3:} \ourModelName performance analysis}
\mytitle{\textbf{Exp 3.(a):} Accuracy in few-samples regime.}
We test our method in the few-training samples regime. 
We train on different sub-sets of the \emph{CIFAR-10} dataset and evaluate on the full test set. 
We compare our \ourModelName-NiN with the baseline NiN and other models using structured filters such as the SRF-NiN \cite{Jacobsen2016CVPR}, 
which also have been shown to generalize well with few training examples. 
\fig{performance_subset_exp} shows the relative accuracy of each model as a percentage of its own top-1 accuracy when trained with 100\% of the data: therefore all models start at 100\% and scores decrease with the decrease in training samples.
We also indicate through the dot size in the plot the relative number of parameters of each model.
Our \ourModelName has the smallest number of parameters. 
This plot shows the expected degradation of the performance of the networks as training data decreases. 
The scores of our \ourModelName-NiN degrade less rapidly, especially when compared to the SRF-NiN and the original NiN model.

\mytitle{\textbf{Exp 3.(b):} Accuracy versus parameter reduction.}
We test the accuracy versus parameter efficiency for our \ourModelName models when compared to a set of baseline CNNs and their SRF versions with fixed scale \cite{Jacobsen2016CVPR} and learned scale \cite{pintea2021resolution}, on \emph{CIFAR-10}, \emph{CIFAR-100}, \emph{STL-10} and \emph{ImageNette} for the ResNet-32 backbone.  
\tab{acc_vs_params} reports accuracies and number of parameters. 
Our \ourModelName layer achieves comparable performance to standard convolutional networks, while reducing the number of parameters 2 to 3 times on NiN and Resnet-32. 
On the EfficientNet-b0 we do not see large parameter reductions because the model heavily relies on 1$\times$1 convolutions which are not replaced with our \ourModelName layers.
On \emph{STL-10} our model with learned $\sigma$ and learned Gaussian derivative order consistently outperforms the other models. 
{On the \emph{ImageNette} dataset our method outperforms the baseline SRF while reducing the number of parameters, as it does not limit the maximum filter frequency.}
The \emph{STL-10} dataset contains high resolution images (96 $\times$ 96 px) allowing for higher frequencies to be present in the data.
While the other methods cannot adapt to varying data frequencies, our models learn this information through the order parameter of the Gaussian derivatives.   
\begin{table}[t]
    \centering
    \resizebox{0.95\columnwidth}{!}{
    \begin{tabular}{@{}llllll@{}}
        \toprule
        \multicolumn{1}{l}{\multirow{2}{*}{}} & \multicolumn{1}{c}{\multirow{2}{*}{{NiN} \cite{lin2013network}} } & \multicolumn{2}{c}{{SRF}} & \multicolumn{2}{c}{{\ourModelName (ours)}} \\
        \multicolumn{1}{l}{} & \multicolumn{1}{c}{} & \multicolumn{1}{c}{{Fixed scale} \cite{Jacobsen2016CVPR}} & \multicolumn{1}{c}{{Learned scale \cite{pintea2021resolution}}} & \multicolumn{1}{c}{{Fixed scale}} & \multicolumn{1}{c}{{Learned scale}} \\ \cmidrule(l){2-2}\cmidrule(l){3-4}\cmidrule(l){5-6}
        {\% Params (count)} & 100\% (0.98M) & 51\% (0.5M) & 53\% (0.52M) & 33\% (0.33M) & 35\% (0.35M) \\
        {CIFAR-10}  & $90.90\%$ & $85.30\%$ & $91.48\%$ & $86.60\%$ & $91.30\%$ \\
        {CIFAR-100} & $67.80\%$ & $61.50\%$ & $68.30\%$ & $61.90\%$ & $67.80\%$ \\
        {STL-10}    & $80.13\%$ & $59.40\%$ & $70.00\%$ & $71.00\%$ & $77.75\%$ \\ \midrule
        \multicolumn{1}{l}{\multirow{2}{*}{}} & \multicolumn{1}{c}{\multirow{2}{*}{{ResNet-32} \cite{he2016deep} }} & \multicolumn{2}{c}{{SRF-ResNet-32}} & \multicolumn{2}{c}{{\ourModelName-ResNet-32 (ours)}} \\
        \multicolumn{1}{l}{} & \multicolumn{1}{c}{} & \multicolumn{1}{c}{{Fixed scale} \cite{Jacobsen2016CVPR}} & \multicolumn{1}{c}{{Learned scale} \cite{pintea2021resolution}} & \multicolumn{1}{c}{{Fixed scale}} & \multicolumn{1}{c}{{Learned scale}} \\ \cmidrule(l){2-2}\cmidrule(l){3-4}\cmidrule(l){5-6}
        {\% Params (count)} & 100\% (0.47M) & 63\% (0.30M) & 65\% (0.31M) & 31\% (0.15M) & 34\% (0.16M) \\
        {CIFAR-10}  & $92.28\%$ & $88.33\%$ & $92.20\%$ & $87.99\%$ & $91.60\%$ \\
        {CIFAR-100} & $67.90\%$ & $65.82\%$ & $67.61\%$ & $63.00\%$ & $67.50\%$ \\
        {STL-10}    & $72.30\%$ & $68.40\%$ & $70.30\%$ & $67.40\%$ & $72.00\%$ \\ 
        {ImageNette} & $86.37\%$ & $78.57\%$ & $81.24\%$ & $80.23\%$ & $83.57\%$ \\ 
        \midrule
        \multicolumn{1}{l}{\multirow{2}{*}{}} & \multicolumn{1}{c}{\multirow{2}{*}{{EfficientNet-b0} \cite{tan2019efficientnet}} } & \multicolumn{2}{c}{{SRF-EfficientNet-b0}} & \multicolumn{2}{c}{{\ourModelName-EfficientNet-b0 (ours)}} \\
        \multicolumn{1}{l}{} & \multicolumn{1}{c}{} & \multicolumn{1}{c}{{Fixed scale} \cite{Jacobsen2016CVPR}  } & \multicolumn{1}{c}{{Learned scale} \cite{pintea2021resolution} } & \multicolumn{1}{c}{{Fixed scale}} & \multicolumn{1}{c}{{Learned scale}} \\ \cmidrule(l){2-2}\cmidrule(l){3-4}\cmidrule(l){5-6} 
        {\% Params (count)} & 100\% (3.6M) & 96\% (3.47M) & 96\% (3.48M) & 95\% (3.43M) & 95\% (3.45M) \\
        {CIFAR-10}  & $92.31\%$ & $89.37\%$ & $93.50\%$ & $84.50\%$ & $90.23\% $ \\
        {CIFAR-100} & $76.20\%$ & $67.50\%$ & $75.81\%$ & $66.89\%$ & $72.50\% $\\
        {STL-10}    & $73.20\%$ & $67.50\%$ & $71.78\%$ & $65.83\%$ & $71.81\% $ \\ \bottomrule
        \end{tabular}
        }
        \caption{\small \textbf{Exp 3.(b):}
        Classification accuracies versus number of parameters on \emph{CIFAR-10}, \emph{CIFAR-100}, \emph{STL-10} and \emph{ImageNette} datasets when comparing the baseline NiN,  Resnet-32 and EfficientNet-b0 with their SRF variants \cite{Jacobsen2016CVPR, pintea2021resolution} and our \ourModelName variants.
        Our method has comparable accuracy with the baselines while largely reducing the number of parameters.
        On the high resolution, encoding more frequencies, \emph{STL-10} dataset our method consistently outperforms the other models.
        }
        \label{tab:acc_vs_params}
        \vspace{-10px}
\end{table}

%% file: discussion.tex
\section{Discussion}



One of the limitations of our model is that computations increase with derivative order, because we rely on the recursive Hermite polynomials to define the Gaussian derivatives.
However, while being computationally more expensive than standard CNNs, we find that \ourModelName models are 25\% faster during training compared to baseline SRF models (time estimates averaged over the complete training epochs) which also rely on the Hermite polynomials. 
The training time speedup comes from only computing 2 Gaussian derivative basis functions per filter.

Another limitation is that the scale learning is fairly unstable and it needs proper regularization and careful learning rate selection.
Additionally, we notice that the orders have the tendency to go towards negative values, requiring clipping during training. 
However, our model greatly reduces the number of parameters when compared to standard 3$\times$3 convolutional layers where instead of learning 9 parameters per kernel, it only needs to learn 3 parameters per kernel: the scale $\sigma$, and the orders $\nu_x$ and $\nu_y$.

%% file: conclusion.tex
\section{Conclusion}
We propose to explicitly learn the frequencies present in the data by encoding these in a trainable network parameter.
We start from the structured filters based on Gaussian derivative basis and make the observation that by learning the order of the Gaussian derivative we can learn to control the filter frequencies. 
We show experimentally that our model can learn the correct frequencies from the data on a synthetic dataset and test the abilities of our model on standard benchmark datasets when compared to NiN, ResNet and EfficientNet backbone architectures. 
Our model degrades gracefully with fewer training samples, and it can achieve good accuracy at large parameter reductions.